\newif\ifarxiv
\arxivtrue 

\ifarxiv
    \documentclass{article}
    \usepackage[margin=1.2in]{geometry}
    \makeatletter
    \newcommand\footnoteref[1]{\protected@xdef\@thefnmark{\ref{#1}}\@footnotemark}
    \makeatother
\else
  \documentclass{bmvc/bmvc2k}
\fi

\usepackage{bm}
\usepackage{enumitem}
\usepackage{siunitx}
\usepackage{tikz}
\usetikzlibrary{calc,arrows.meta,positioning,decorations.markings}
\usepackage{amsfonts}
\usepackage{amsmath}

\usepackage[linesnumbered,ruled]{algorithm2e}

\SetCommentSty{mycommfont}

\usepackage{microtype}
\usepackage{graphicx,calc}
\usepackage{subfigure}
\usepackage{booktabs} 

\usepackage{hyperref}

\newlength\myheight
\newlength\mydepth
\settototalheight\myheight{Xygp}
\newcommand*\inlinegraphics[1]{%
  \settototalheight\myheight{Xygp}%
  \settodepth\mydepth{Xygp}%
  \raisebox{-\mydepth}{\includegraphics[height=\myheight]{#1}}%
}

\newcommand\ndash{\mathop{\mbox{$n$-}}}
\newcommand\dashw{\mathop{\mbox{-$\mathsf{w}$}}}
\newcommand{\TMS}{{\mkern-1.5mu\times\mkern-1.5mu}}

\newenvironment{tightitemize}
{ \begin{itemize}
    \setlength{\itemsep}{4pt}
    \setlength{\parskip}{0pt}
    \setlength{\parsep}{0pt}     }
{ \end{itemize}                  }

\usepackage{mathtools}

\DeclarePairedDelimiter\floor{\lfloor}{\rfloor}

\begin{document}


\ifarxiv
	\title{A New Benchmark and Progress Toward Improved Weakly Supervised Learning}
	\author{Jason Ramapuram\footnote{Equal contributions.}
          \footnote{University of Geneva \& University of Applied
            Sciences, Western
          Switzerland}\ , jason@ramapuram.net \\ Russ Webb\textsuperscript{*}\footnote{Apple Inc, Cupertino, CA}\ , rwebb@apple.com}
	\maketitle
\else
        \title{A New Benchmark and Progress Toward Improved Weakly Supervised Learning}
        \addauthor{Jason Ramapuram}{jason@ramapuram.net}{2}
        \addauthor{Russ Webb}{rwebb@apple.com}{1}

        \addinstitution{
          Apple Inc\\
          Cupertino, California, USA
        }
        \addinstitution{
          University of Geneva \& \\University of Applied Sciences, \\
          Western Switzerland
        }

        \runninghead{Ramapuram, Webb}{Toward Improved Weakly Supervised Learning}

        \maketitle
\fi





\begin{abstract}
{\em Knowledge Matters: Importance of Prior Information for Optimization} \cite{gulccehre2016knowledge}, by G\"{u}l\c{c}ehre et. al.,
sought to establish the limits of current black-box, deep learning
techniques by posing problems which are difficult to learn without engineering
knowledge into the model or training procedure.  In our work, we
solve the previous {\em Knowledge Matters} problem with 100\% accuracy
using a generic model, pose a more
difficult and scalable problem, All-Pairs, and advance this new problem
by introducing a new learned, spatially-varying histogram model called TypeNet which outperforms conventional
models on the problem.  We present results on All-Pairs where our model
achieves 100\% test accuracy while the best ResNet models achieve 79\%
accuracy.  In addition, our model is more than an order of magnitude smaller than Resnet-34.  The challenge of solving larger-scale
All-Pairs problems with high accuracy is presented to the community for investigation.
\end{abstract}

\section{Introduction}
Deep neural networks are powerful functional approximators, allowing for
the learning of complex tasks that were not solvable by traditional
machine learning methods. Recently, \cite{gulccehre2016knowledge}
suggested that there exist problems that neural networks would not be
able to solve without the guidance of human insight; they define and
study the Pentomino problem as an example of this class of problems. 
For the Pentomino problem, we demonstrate that extra
knowledge is not necessary by solving the problem with a small, deep neural network (DNN).
Having found a solution to Pentomino, we introduce a new, scalable problem and present progress toward its solution.

To understand the limits of weakly supervised learning applied to generic models, we divide the task of solving a problem into the application of known
techniques and the engineering of the system (the model plus training data and proceedure).  The palette of known
techniques is constantly improving and is what enables solving the
Pentomino problem with current techniques.  The incorporation of explicit
knowledge into an engineered solution can be estimated by how many problem
specifics can be inferred/discovered by an inspection of the model architecture and the training
procedure.  Common deep-learning techniques are a codification of
knowledge into reusable components which require minimal insight to
select.  For instance, batch norm \cite{ioffe2015batch} speeds
convergence and reduces hyper-parameter sensitivity, jump connections
\cite{srivastava2015training, he2016deep} enable deeper models,
convolutions \cite{lecun1998gradient,krizhevsky2012imagenet} are useful
in spatially-invariant vision problems, and sparse activations
\cite{srivastava2014dropout, wan2013regularization} reduce overfitting
by restricting the flow of information.  Even simple observations, like these,
allow the practitioner to select components suitable for the problem. 
These techniques are all excellent examples of knowledge refined into
heuristically selectable, generic techniques.

We lack ready-to-apply techniques for some problems and much of the
research in the field moves us toward more turn-key application of
learning algorithms; for example, the latest AlphaGo
\cite{silver2017mastering} is trained without expert human examples. 
Some examples of the work done to engineer problems with human knowledge
are engineering model sub-components to include problem details and 
adding sub-goal labels or objective functions.  Successful engineering of a 
solution for a particular problem can lead to either a specific solution only 
applicable to the problem studied or, more usefully, to broadly reusable 
techniques or insights.  The later outcome is our goal in presenting the following 
contributions:
\begin{enumerate}
  \item demonstration of solving the Pentomino problem from {\em Knowledge Matters} \cite{gulccehre2016knowledge} with conventional techniques (both model and training)
  \item new, scalable challenge problem, All-Pairs, with (effectively) infinite data \cite{allpairs2018}
  \item sampling of existing techniques' performance on All-Pairs as baselines
  \item new, generic model, TypeNet \cite{allpairs2018}, which out-performs the baselines on All-Pairs.
\end{enumerate}
The All-Pairs dataset generator and TypeNet reference code are available at \href{https://github.com/apple/ml-all-pairs}{https://github.com/apple/ml-all-pairs}.

\section{Related Work} \label{related}

Our work spans two distinct areas of machine learning: learning under
weak supervision and extracting
relational information from high-dimensional data. By \emph{weak supervision}, we mean that our
model is required to solve a high-level task such as the binary
classification proposed on the left of Figure \ref{pento_sprites}
by observing only raw pixels.  
\ifarxiv
The information content of the gradients relative to sampling noise has been
studied \cite{2017failures} as a way of characterizing the difficulty of end-to-end learning.
\fi

Prior work in weak supervised learning (WSL) in the image domain has
focused on image segmentation by classifying them with a standard
multi-class loss objective \cite{oquab2015object} or by utilizing an
alternate loss such as a score-based \cite{durand2016weldon} objective.
Unsupervised representation learning can also be used to aid the model
in learning the end objective. The recent work of {\em Learning to Count}
\cite{noroozi2017representation} proposed a method for representation
learning in an unsupervised setting by using a pre-trained network
to learn counting of visual primitives. This method works well
when the features extracted from the pre-trained network are
semantically relevant to the current learning objective. Our work
differs from this and the WSL objective in the amount of supervision
provided to the model. We focus on supervised tasks where a model is not
provided with sub-problem class labels (or any other structured,
supervised information) and needs to learn a high-level representation of
the visual scene using few binary labels for each whole image.

Extracting relational information with neural networks has been studied in many settings
from text-based relationships \cite{das2016chains, weston2014memory} to
visual query answer (VQA) models such as the recent work of Relational
Networks \cite{santoro2017simple} and Show-Tell-Attend
\cite{xu2015show}. Relational Networks have been used to learn
relationships between objects in a scene given a rich textual query,
such as the CLEVR dataset \cite{johnson2017clevr}, which provides input
in the form of an image coupled with a textual query.  Despite having a pair-wise structure that we
intuitively think is useful for our All-Pairs problem, a Relational
Network \cite{santoro2017simple} does not solve our proposed dataset
(Section \ref{all_pairs}). To solve our problem, we introduce a model
called TypeNet which aggregates channel-wise statistics and solves the
overall task by combining these statistics.

Our TypeNet model takes inspiration from winner-take-all (WTA)
strategies \cite{srivastava2013compete,maass2000computational} and can build a set of problem-related,
local statistics to combine for predicting the end
objective. We demonstrate empirically that TypeNet outperforms
state-of-the-art models such as ResNet18, Resnet34 \cite{he2016deep},
VGG19 \& VGG16 with batch norm \cite{simonyan2014very}, and InceptionV3
\cite{szegedy2016rethinking} on the proposed All-Pairs problem.

\section{Solving the Pentomino Problem}

\begin{figure*}[!htb]
\minipage[t]{0.468\textwidth}
  \includegraphics[width=\linewidth]{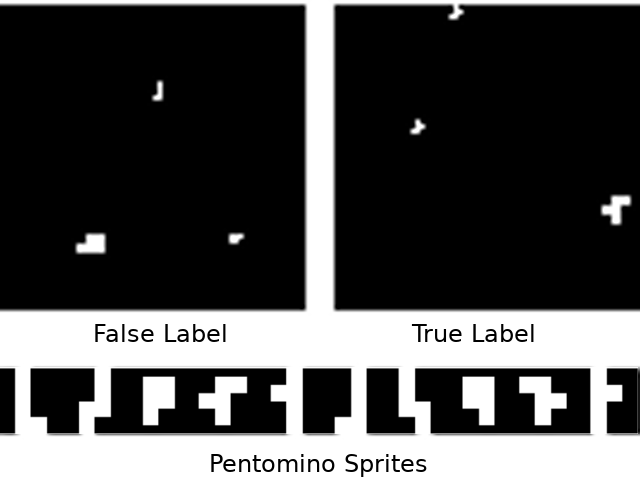}
  \endminipage\hfill
\minipage[t]{0.532\textwidth}
  \includegraphics[width=\linewidth]{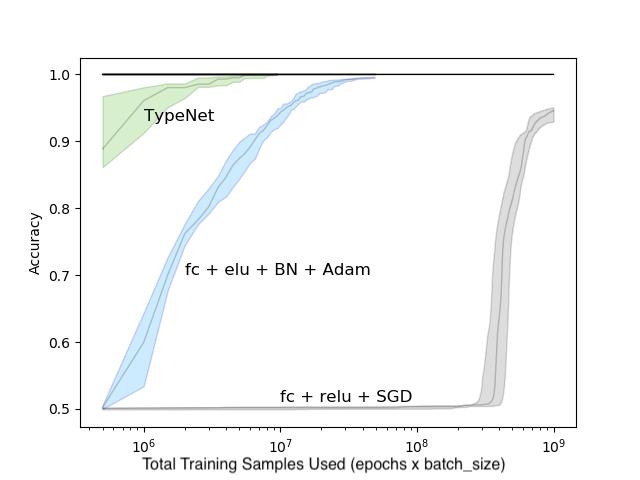}
\endminipage\hfill
\caption{\emph{Left}: The Pentomino sprites and two examples illustrating the $true$ and $false$ classes.
  \emph{Right}: Test accuracy (median and inner quartiles, 10 trials) on the Pentomino problem with and without modern training advances. Note, log-scale of x-axis.}
\label{pento_sprites}
\end{figure*}

{\em Knowledge Matters} \cite{gulccehre2016knowledge} explores the extent to
which neural networks are able to learn problems given minimal
supervised information. Their formulation has a fully defined loss
function; however, the gradient of the loss with respect to the
parameters provides no direct information about potentially useful
subtasks such as segmentation, object classification, or counting. They concluded that the networks and training methods they tested converged to a
local minima.

The {\em Knowledge Matters} demonstration utilized the Pentomino dataset, which
is formed from a set of sprites \cite{gulcehre_2015}
shown in Figure \ref{pento_sprites}. The
dataset is generated by placing three sprites onto a canvas $C \in
\mathbb{R}^{64\TMS64}$. Each sprite undergoes a random rotation 
(\ang{0}, \ang{90}, \ang{180}, or \ang{270}) and integer scaling ($1\times$ or $2\times$).  The goal of the
neural network is to predict a 1 if the rotated and scaled sprites in an
image are the same and 0 otherwise. One possible solution to the
Pentomino problem is to learn to segment, classify, and count the number
of underlying objects in the image. The challenge (claimed impossible in \cite{gulcehre_2015})
is to find a solution using a generic network given only the binary
label for each image.

G\"{u}l\c{c}ehre et. al \cite{gulccehre2016knowledge} observed that
``black-box machine  learning  algorithms  could  not  perform  better 
than  chance on [the Pentomino problem].'' Decomposing
the problem into two stages however, made the task easily solvable. The
first stage in the decomposition was a classification
step, where extra label information was provided to the model. Given the
predicted classes, the second stage projected this output to the Bernouili
log-likelihood objective. Using some of the recent advances in 
DNN training, we are able to completely solve the original
problem demonstrated in {\em Knowledge Matters}; we do so without the
requirement of an intermediary model or the addition of extra
information. We also experimented with a reproduction of the model 
proposed in the paper and found that given enough time (over 1000
epochs) the model does make progress on the Pentomino problem, as shown
in Figure \ref{pento_sprites} in gray. This observation is in line with
recent insights of \cite{hoffer2017train} that discuss the effects of
training duration and batch size.

The fully-connected (fc) model presented in \cite{gulccehre2016knowledge} was composed of layer sizes
[2050, 11, 1024] and trained with ADADelta \cite{zeiler2012adadelta} and weight regularization. The 11-unit layer
served as a bottleneck to bring structural information into the
network.  We leverage four recent advances to solve the Pentomino problem: Batch
Normalization (BN) \cite{ioffe2015batch}, Exponential Linear Units
\cite{clevert2015fast}, the Adam optimizer \cite{kingma2014adam}, and
Xavier initializations \cite{glorot2010understanding}. In constrast to
the large model employed in \cite{gulccehre2016knowledge}, we use a
fully-connected network with layer sizing of
$[32, 64, 12, 32, 8]$; this translates to a 98.5\% reduction of the
total number of model parameters. Comparable in size to the largest
training sets used in \cite{gulccehre2016knowledge}, 486k
samples were used for training and 54k samples were held out for testing.

G\"{u}l\c{c}ehre et. al \cite{gulccehre2016knowledge} were only able to
train black-box (generic), fully-connected models to achieve 50\% accuracy on the
Pentomino dataset. Their best model, after significant
hyper-parameter search, resulted in a 5.3\% training and 6.7\% test error
on the 80k Pentomino training dataset. This performance was achieved 
via a two-stage network that induced structural information
into the neural network. On the same training set, we achieved a 1\% error using a black-box neural network with the
5-layer network described above.  Figure
\ref{pento_sprites} shows the training accuracy for the original
{\em Knowledge Matters} network (gray), our modification (blue), and our TypeNet model (green, see Section \ref{typenetdetails}) on
the Pentomino problem (note the log scale on the x-axis).


\section{The All-Pairs Problem} \label{all_pairs}
\begin{figure}[h!]
\vskip 0.2in
\begin{center}
\scalebox{0.6}{\centerline{\includegraphics[width=\columnwidth]{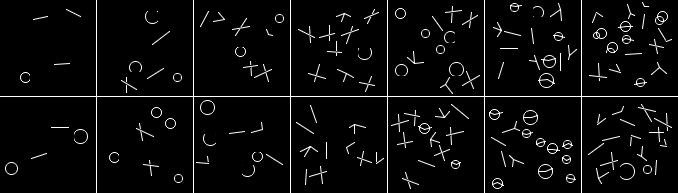}}}
\caption{All-Pairs examples from 2-2 on the left to 8-8 on the right.  The bottom row is $true$ and the top row is $false$.}
\label{example8}
\end{center}
\vskip -0.2in
\end{figure}

\subsection{Definition and Examples}

Extending the ideas in the Pentomino problem, we use anti-aliased white
symbols on a black background to construct the following new problem.
The \textit{N-K} All-Pairs problem contains \textit{2N} symbols from an
alphabet of \textit{K} choices.  Each example is \textit{true} if each of its
symbols pairs with a symbol of the same type without reuse,
and \textit{false} otherwise.  Symbols are positioned randomly with no
overlap.  Symbols are of similar scales, ranging from
10--18 pixels across, and have differing symmetries (for instance, some
are rotationally invariant, while others are not).  The exact structure
and variations of each symbol are given by the generator code supplied
online \cite{allpairs2018}.

Each symbol is shown below with the
number of unique ways it can appear, as configured in our experiments.  In contrast, the
Pentomino problem used 8
variations for each symbol. The symbols are used in the order given, so the 4-4 All-Pairs problem will use \textbf{circle}, \textbf{line}, \textbf{cross}, and \textbf{angle}.  For this work a $76\TMS76$ image is used for $N<6$ and a
$96\TMS96$ image is used for larger $N$.

\begin{table}[htp]
  \begin{center}
\scalebox{0.9}{
\begin{tabular}{r r l r r r l r}
id & name & examples & cardinality &id & name & examples & cardinality \\
\hline
1 & \textbf{circle}        & \inlinegraphics{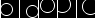}  & 165   & 10 & \textbf{box}          & \inlinegraphics{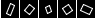}  & 480   \\
2 & \textbf{line}          & \inlinegraphics{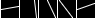}  & 174   & 11 & \textbf{box-diagonal} & \inlinegraphics{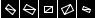} & 518   \\
3 & \textbf{cross}         & \inlinegraphics{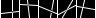}  & 45.3k & 12 & \textbf{barbell}      & \inlinegraphics{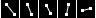} & 78    \\
4 & \textbf{angle}         & \inlinegraphics{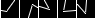}  & 39k   & 13 & \textbf{dot-line}     & \inlinegraphics{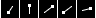} & 156   \\
5 & \textbf{3-star}        & \inlinegraphics{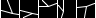}  & 1.43M & 14 & \textbf{z}            & \inlinegraphics{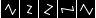} & 518   \\
6 & \textbf{theta}         & \inlinegraphics{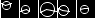}  & 20k   & 15 & \textbf{triangle-lid} & \inlinegraphics{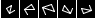} & 1036  \\
7 & \textbf{phi}           & \inlinegraphics{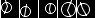}  & 20k   & 16 & \textbf{dot-mid-line} & \inlinegraphics{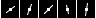} & 78    \\
8 & \textbf{2-circle}      & \inlinegraphics{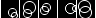}  & 7k& 17 & \textbf{hourglass}    & \inlinegraphics{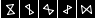} & 518   \\
9 & \textbf{circle-3star}  & \inlinegraphics{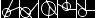}  & 7.15M& 18 & \textbf{triangle}     & \inlinegraphics{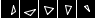} & 11.8k \\
\end{tabular}
}
\end{center}
\label{default}
\end{table}
\vskip -0.2in

\noindent
Figure \ref{example8} shows a \textit{true} and a \textit{false} example
for the 2-2 to 8-8 All-Pairs problem.  A data generator for All-Pairs is
used to generate on-demand, unique training examples (the 4-4
All-Pairs problem has approximately $10^{28}$ unique images), and a fixed
validation set is generated at the start of training.  The separability of the eighteen
symbols was confirmed by training a simple conv-net to 100\% test accuracy in 350k training samples.

\subsection{Comparison with Conventional Results}
Conventional algorithms from the literature have difficulty with the 4-4
All-Pairs problem, as shown in the following table.  Clearly, of the
hundreds of conventional, valuable DNN algorithms, there may exist some
that can solve the 4-4 problem.  One open challenge is to identify them
and extend training techniques to efficiently solve these types of problems.
Of the runs of each algorithm summarized below, none achieved more than
92\% test accuracy after training on 100M samples.  An expert human made one mistake in 100 samples for each of the All-Pairs problem from difficulty 4-4 to 7-7, taking 8-9 seconds to classify each image.  Humans use sequential attention and working memory to do the All-Pairs task, suggesting the task as a benchmark for building sequential models.
TypeNet consistently achieves 100\% test accuracy in the 4-4 All-Pairs
problem using 20k test samples.  

\begin{center}
\begin{tabular}{ c c c c c }
 algorithm & model size & normalized size & accuracy & std deviation \\
\hline
 TypeNet [$\times$10] & 918k & 1.0 & \textbf{1.000} & \textbf{0.000} \\
 Expert Human [$\times$1] & -- & -- & \textbf{0.990} & -- \\
 Relational Net [$\times$10] & \textbf{630k} & \textbf{0.7} & 0.867 & 0.078 \\
 \ifarxiv
 ConvNet (\S\ref{comptocnn}) [$\times$4] & 9.9M & 11 & 0.808 & 0.093    \\
 \fi
 Inception v3 [$\times$10] & 22M & 24 & 0.803 & 0.079    \\
 Resnet-34 [$\times$10] & 21M & 23 & 0.788 & 0.068   \\
 Resnet-18 [$\times$10] & 11M & 12 & 0.711 & 0.157    \\
 Vgg19 [$\times$6] & 139M & 151 & 0.509 & 0.002    \\
 Vgg16 [$\times$3] & 134M & 146 & 0.506 & 0.002    \\
\end{tabular}
\end{center}

\section{Toward an All-Pairs Solution}

\begin{algorithm}
\label{tn_algo_block}
\caption{TypeNet algorithm}
\SetAlgoNoEnd
\SetKwComment{tcp}{\# }{}
    \KwData{ 
    	\begin{itemize}
		\renewcommand\labelitemi{--}
		\setlength\itemsep{-0.2em}
		\item Number of layers, $N_c$ and $N_f$.
		\item Number of type branches, $N_t$, and spatial branches, $N_s$.
		\item Activations, \textsc{Ac}, and convolutions, \textsc{Conv}, for feature extraction layers.
		\item Activations, \textsc{A}, and $n$ 1x1 convolutions, \textsc{Conv1$\TMS$1}, for type matching.
		\item Spatial diversity operations, \textsc{Spatial}.
		\item Activations, \textsc{Afc}, weights, $W$, and biases, $B$, for fully-connected layers.
	\end{itemize}
    }

$C = \textsc{Image}$ \tcp*[f]{convolution block} \\
\For{$i=[1 \to N_c)$}{
	$C = \textsc{Ac}_i(\textsc{Conv}_i(C))$ \\
	$C = \textsc{BatchNorm}(C)$
}
\BlankLine
T = $\sum_{i=0}^{N_t}{\textsc{A}_i(\textsc{Conv1$\TMS$1}_i(C))}$
\BlankLine
Y = \textsc{Concatenate}\Big(\big[ $\sum_{w,h}{\textsc{Spatial}_i(T)}$ for $i = [0 \to N_s)$ \big]\Big)

\BlankLine
\For(\tcp*[f]{fully-connected layers}){$i=[0 \to N_f)$} { 
	$Y = \textsc{Afc}_i(W_i Y + B_i)$ \\
	$Y = \textsc{BatchNorm}(Y)$
}

\KwRet{$\textsc{SoftMax}(Y)$}
\end{algorithm}


%
%
%

\subsection{Type-Net Model} \label{typenetdetails}

After verifying that a fully-connected model can easily solve the 4-4 All-Pairs
problem from the histogram of symbols in each
image, we designed and tested a generic model capable of
learning a similar, whole-image statistic.  The resulting model was created
using insights derived from the All-Pairs problem, but does not make use of
explicit problem details or enhanced training data.

We refer to the resulting network as a TypeNet because it estimates the affinity
of each receptive field to $n$ ideal types (via a dot-product) and then
aggregates those type-affinities over the spatial extent.  This spatial
summation is global for solving the All-Pairs problem, but could
be spatially restricted to produced learned features similar to
histogram of gradients (HOG) found in \cite{mcconnell1986method}.  A learned attention mask could also generalize the summation to salient areas of each image.

Model details can be found in the supplementary material and in the online sample code found at \cite{allpairs2018}.  The general algorithm for TypeNet is presented in Algorithm \ref{tn_algo_block}.  The algorithm begins and ends conventionally with a convolution stack and fully-connected layers, respectively.  Lines 5 and 6 show the key steps for the algorithm:
\begin{itemize}
\item line 5, the $1\TMS$1 convolution implements a dot-product similarity with a learned kernel, these are the ``types'' of TypeNet.
\item line 5, the activation, \textsc{A}$_i$, applied was experimentally studied:
	\begin{itemize}
	\item \textsc{A}$_i$ = \textsc{SoftMax} in the feature dimension, gives a soft $N_t$-hot representation here which was seen to reduce variance in training times).
	\item \textsc{A}$_i$ = \textsc{Identity} was the most versatile activation and can be seen as creating a ``type'' difference operator
	\end{itemize}
\item line 5, superposition (via summation) of learned template matching
\item line 6, diversify spatially with non-linear operators such as \textsc{MaxPool}.
\end{itemize}

The goal of introducing TypeNet is to expand the palette of
techniques available to solve similar types of problems and decrease
the problem specific reasoning required in similar
domains (such as parity, counting, holistic scene understanding, and
visual query answer), which can be solved from a histogram-like summary of
local statistics.

\subsection{Contrast to Relational Methods}

Relational neural learning generally accomplishes it's goal by
learning a functional over $(i,j)$ tuples in a latent feature space $f$. In
Relational Networks \cite{santoro2017simple} for example, the
model learns two functionals $[h, g]$ (parameterized by
deep-neural networks) that \textbf{exhaustively} operate over \textbf{all} $(i,j)$ pairs
in the latent feature space of a deep-convolutional network as shown
in the table below.
Memory Networks \cite{weston2014memory} on the other hand learn a probabilistic relationship
between the input query (embedded into a feature representation) $f_i$ and an
associated set of memory vectors $M = \{m_1, ... m_i, m_N \}$, followed by a smoothed weighting against an embedded query
vector $c_i$.
\vskip -0.2in
{\renewcommand{\arraystretch}{1.15}
\begin{center}
\begin{tabular}{ c | c } \hline
  Relational Networks & Memory Networks \\ \hline
  $g(\sum_i \sum_j h(f_i, f_j))$ & $p_i = \text{softmax}(f_i^T,
                                         m_i) \hspace{0.2in}    o_i =
                                         \sum_i p_i c_i$
\end{tabular}\label{rel_mem_net}
\end{center}
}

Relational Networks \cite{santoro2017simple} have high computational
complexity when the dimensionality of the feature-space
$f$ is large. Memory-networks on the other hand scale proportionate
to the number of embedded memories dim($M$). Our objective with TypeNet
is twofold: relax computational constraints compared to these relational models and incorporate the probabilistic smoothing of Memory Networks \cite{weston2014memory}.

We reduce the computational complexity by forcing the model to divide
the input representation through a set of $N_t$ branches. This
division allows the model to learn a disparate feature
representation per branch. Rather than learning over every $(i,j)$ as in Relational Networks \cite{santoro2017simple} we
approximate this with a spatial sum after our branch-divide
strategy.

During our branching strategy we do a sum across an activated feature
space; this can be interpreted as a probabilistic weighting of the
features of each individual branch against each other. Training TypeNet to convergence is \textbf{8-10 times faster} than training a comparable Relational Network and produces \textbf{more accurate results} in the
weak-supervised learning scenario of All-Pairs.

\subsection{All-Pairs Result} \label{allpairsresult}

\begin{figure*}[!htb]
  \includegraphics[width=\linewidth]{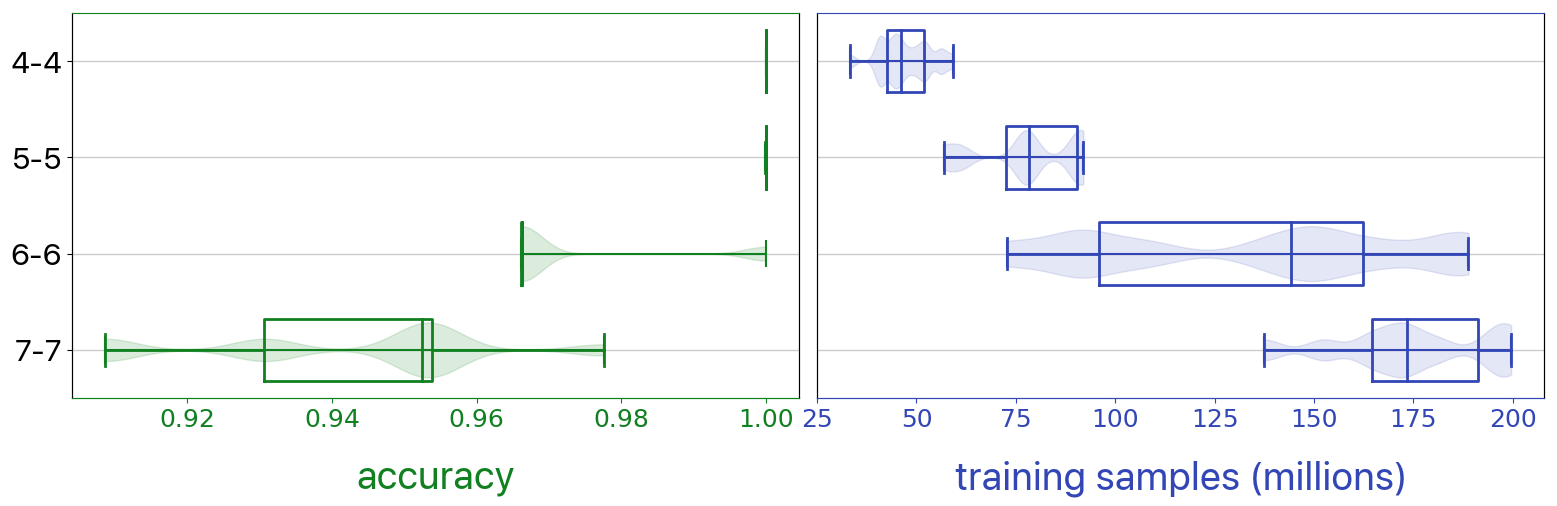}
\caption{Training results, showing validation accuracy and total number of training sample for TypeNet on increasingly difficult versions of All-Pairs, from 4-4 to 7-7.  Shading shows the distribution over 10 trials.  Note, conventional DNN models cannot solve the 4-4 problem.}
\label{solving_all_pairs}
\end{figure*}

As described, the TypeNet for the All-Pairs problem has 1.04M trainable parameters
(many times smaller than the baseline models).
Unless otherwise noted, we used the following training setup for the
TypeNet results on the All-Pairs problem: 4 GPUs, batch size 600, Adam
with learning rate = 0.001 and no weight decay, cross-entropy loss, test
results reported every 50k training samples, and 100M total training samples.
A 100M sample training run typically takes 20 hours on 4$\TMS$P100 GPUs.

The main hyper-parameters and architecture-variations explored are the
feature activation, number of branches ($k$), and number of features
($n$).  Details of those studies can be found in the supplementary material.  We concluded that $k = 2$ and $n = 64$ performed well on the All-Pairs problem.  Increasing the number of features to 96 results in slightly lower training times, at a cost of a larger model.  All options explored reached 100\% accuracy.

The TypeNet approach cannot easily solve every All-Pairs
problem; Figure \ref{solving_all_pairs} shows results for the 4-4 to 7-7 All-Pairs problem.
We see an increase in the magnitude and variance of the number of samples needed for convergence.  The plot shows the results of 10 training runs for each difficulty level; TypeNet can solve the first 3 of these challenges to 100\% validation accuracy.  No model and training methodology has been found that solves the 7-7 problem to 100\% accuracy.  An inspection of the errors made by the best 7-7 solution shows that they are systematic, unambiguous errors.

\begin{figure*}[!htb]
\centering
\minipage{0.6\textwidth}
  \includegraphics[width=\linewidth]{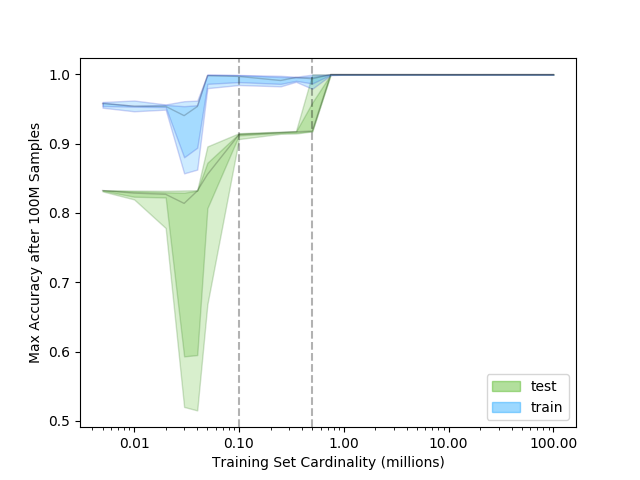}
\endminipage\hfill
\caption{Training the TypeNet on 4-4 All Pairs with 100M samples drawn from a fixed-size training set.}
\label{data_setsize_fig}
\end{figure*}

\subsection{Training Set Size}

In Figure \ref{data_setsize_fig}, we show the effect of reducing the
cardinality of the training data from effectively infinite to sizes
smaller than the total number of training samples presented.  A training
set cardinality of 100k is minimal for 90\% test accuracy and 500k is
minimal for some trials to reach 100\% test accuracy.  The increased variance in both train and test accuracy at
cardinality 30k is interesting.  We hypothesize this is due to sampling
noise for these small sizes leading to significantly different train and
test distributions.  For larger cardinality, both sets consistently
represent the same distribution; for smaller sets, learning is
limited enough that distribution differences are not apparent.

To avoid the overhead of datasets on disk, varying the training set
cardinality is accomplished using an array of seeds for the data
generator.  Each seed is used to generate 1k samples.  When each seed in
the list has been used once, the list is shuffled and the process
starts back at the beginning of the list.

\subsection{Other Applications}

TypeNet was evaluated on other datasets to determine its applicability
to common classification problems.  The following table presents results for the test
accuracy from four training runs.  For training, each dataset was augmented by random original-size
crops (padding of 4), random rotations from \ang{0} to \ang{4}, and normalized by subtracting 0.5.
CIFAR10 and Fashion MNIST \cite{fashionref}
were also augmented with random horizontal flips.  A detailed discussion and comparison with a simple convolutional net can be found in Supplement \ref{comptocnn}.

\begin{center}
\begin{tabular}{ r | c c | c c  }
              & \multicolumn{2}{c|}{ConvNet}                        & \multicolumn{2}{c}{TypeNet}  \\
dataset  & accuracy & \# parameters & accuracy & \# parameters \\
\hline
MNIST               & 0.9953 $\pm$ 0.0002    & 2M & \textbf{0.9971 $\pm$ 0.0006} & \textbf{1M}  \\
Fashion-MNIST & \textbf{0.9409 $\pm$ 0.0005}    & 2M & 0.9346 $\pm$ 0.0011 & \textbf{1M} \\
CIFAR10           & 0.7773 $\pm$ 0.0013     & 2.5M & \textbf{0.8820 $\pm$ 0.0080} & \textbf{1M} \\
4-4 All-Pairs      & 0.8080 $\pm$ 0.0925     & 9.9M & \textbf{1.0000 $\pm$ 0.0000} & \textbf{1M} \\
\end{tabular}
\end{center}


For these classification
tasks, adding more spatial information via two parallel pathways branching
from the \textbf{similarity} step (algorithm line 5) and joining at the \textbf{concatenation} step (line 6)
was useful.  One of these extra pathways has a \textsc{MaxPool3x3} and the other has \textsc{AvgPool3x3} after
the \textbf{similarity} step.  This enhanced model also solves the All-Pairs problem and has 10\% more parameters than the simpler TypeNet presented as a minimal version for All-Pairs.

\section{Conclusion}

In this work, conventional training methods and model features have been
demonstrated to solve a previously unsolved task by training a black-box
model to solve the Pentomino problem.  The All-Pairs problem is
introduced as a challenge to the research community by measuring the
limits of conventional model performance, introducing a model
advancement (TypeNet) to solve such problems, and measuring the limits
of TypeNet on the All-Pairs and conventional image classification
benchmarks.

The following extensions to the TypeNet model and its training may
prove useful or generate further insights: (1) filtering the data
generator output to study supervised and unsupervised curriculum
learning, (2) generating multi-scale statistics before
the final \textbf{fully-connected} layers, (3) annealing a $softmax$-type
activation during training to help the network seek better minima,
and (4) using the TypeNet structure in a residual architecture by adding the
post-\textbf{superposition} block of features back to the \textbf{conv}
block.  The hope is to direct research toward valuable investigations
and to promote a methodology of falsifiable scientific claims both by
falsifying previous claims and by making further claims which, if we believe
Popper \cite{popper59}, are likely to be false.

\ifarxiv
    \bibliographystyle{abbrv}
\fi

\bibliography{0_Progress_Toward_Improved_Weakly_Supervised_Learning}

\newpage
\section{Supplement}

\subsection{TypeNet Configuration}
Figure \ref{typenet_fig_sup} shows the data flow for our model as configured for the All-Pairs problem.  The Tables \ref{global_config}, \ref{detail_config_1}, and \ref{detail_config_2} present the detailed network configuration (also found in the sample code distributed with the dataset generator):

\begin{figure*}[!htb]
  \scalebox{0.9}{
\begin{tikzpicture}
\begin{scope}[
  thick,decoration={
        markings,
      }
    ]
\node[text={rgb:red,4;green,2;yellow,1}] (image) at (0,0) {image};
\node[text=blue] (conv) at (1.5, 0) {conv};
\draw[postaction={decorate}] (image.east) -- (conv.west);

\node[text=blue] (a11) at (3, -0.5) {$1\TMS1_2$};
\node[text=blue] (b11) at (3, 0.5) {$1\TMS1_1$};
\draw[postaction={decorate}] (conv.east) -- (a11.west);
\draw[postaction={decorate}] (conv.east) -- (b11.west);

\node (sm_3) at (5.0, -0.5) {\textsc{identity}};
\node (sm_4) at (5.0,  0.5) {\textsc{identity}};
\draw[postaction={decorate}] (a11.east) -- (sm_3.west);
\draw[postaction={decorate}] (b11.east) -- (sm_4.west);

\node (p_2) at (7,  0) {$+$};
\draw[postaction={decorate}] (sm_3.east) -- (p_2.west);
\draw[postaction={decorate}] (sm_4.east) -- (p_2.west);

\node (mp3) at (9.5, 1) {\textsc{MaxPool(3)}};
\node (mp5) at (9.5, -1) {\textsc{MaxPool(5)}};
\draw[postaction={decorate}] (p_2.east) -- (mp3.west);
\draw[postaction={decorate}] (p_2.east) -- (mp5.west);

\node (h_1) at (11.5,  0)  [label={[xshift=0.0cm, yshift=-0.7cm]$\sum\limits_{w,h}$}] {\ \ \ \ };
\node (h_2) at (11.5,  -1)  [label={[xshift=0.0cm, yshift=-0.7cm]$\sum\limits_{w,h}$}] {\ \ \ \ };
\node (h_3) at (11.5,  1)  [label={[xshift=0.0cm, yshift=-0.7cm]$\sum\limits_{w,h}$}] {\ \ \ \ };

\node (num) at (12.7,  0.85) {$[n]$};

\draw[postaction={decorate}] (p_2.east) -- (h_1.west);
\draw[postaction={decorate}] (mp3.east) -- (h_3.west);
\draw[postaction={decorate}] (mp5.east) -- (h_2.west);

\node[text=blue] (fc) at (13.5, 0) {$fc$};
\draw[postaction={decorate}] (h_1.east) -- (fc.west);
\draw[postaction={decorate}]  (h_2.east) -- ([yshift=-4pt] fc.west);
\draw[postaction={decorate}]  (h_3.east) -- ([yshift=4pt] fc.west);


\end{scope}
\end{tikzpicture}
}
\caption{Model data-flow used for All-Pairs.}
\label{typenet_fig_sup}
\end{figure*}
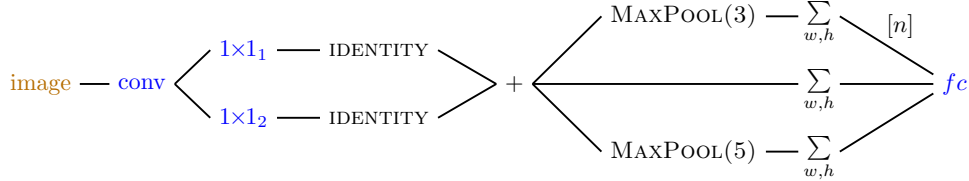

\begin{table}[!htb]
  \centering%
    \begin{tabular}{ l l }
    parameter & value \\
    \hline
    $N_c$ & 4  \\
    $N_f$ & 4  \\
    $N_t$ & 2  \\
    $N_s$ & 3  \\
    $n$ & 64  \\
    \textsc{Ac} & \textsc{Elu}  \\
    \textsc{A} & \textsc{Identity}  \\
    \textsc{Spatial} & \{\textsc{Identity}, \textsc{MaxPool3x3}, \textsc{MaxPool5x5}\}  \\
    \textsc{Afc} & \textsc{Elu}  \\
    \end{tabular}
  \caption{Model level parameters for TypeNet as used to solve All-Pairs.}\label{global_config}
\end{table}

\noindent
For the activation, \textsc{A}$_i$, the most useful activations were found to be \textsc{Identity}, \textsc{Selu}, and
\textsc{SoftMax}.  \textsc{SoftMax} is in the feature, rather than spatial
dimension.  Via architecture search, \textsc{Identity} is the most generally useful activation, though \textsc{SoftMax} tended to reduce training times (probably because it forms a strong approximately-sparse bottleneck).  

The convolution block used \textsc{Elu} activation, no bias, batch norm (post-activation),
and padding to align the convolution filters with the image edges.  It's layer-wise characteristics are detailed below.  For larger images, a stride of 2 was used in \textsc{Conv}$_3$.
\begin{table}[!htb]
  \centering%
    \begin{tabular}{ l l }
    parameter & features, size, stride  \\
    \hline
    \textsc{Conv}$_1$ & 128, 3, 1  \\ 
    \textsc{Conv}$_2$ & 128, 5, 2  \\
    \textsc{Conv}$_3$ & 128, 5, 1  \\
    \textsc{Conv}$_4$ & 128, 3, 1  \\
    \end{tabular}
  \caption{Convolution block parameters for TypeNet as used to solve All-Pairs.}\label{detail_config_1}
\end{table}

\noindent
The fully-connected layers have input of size $m=N_s \TMS n$ and their configuration is detailed below:
\begin{table}[!htb]
  \centering%
    {\renewcommand{\arraystretch}{1.15}
    \begin{tabular}{ l l }
    parameter & value  \\
    \hline
    $fc_1$ & $m$-\textsc{Elu}-bnorm  \\
    $fc_2$ & $\floor*{\frac{m}{2}}$-\textsc{Elu}-bnorm  \\
    $fc_3$ & $\floor*{\frac{m}{4}}$-\textsc{Elu}-bnorm  \\
    $fc_4$ & $2$-\textsc{Identity}  \\
    \end{tabular}
    }
  \caption{Fully-connected parameters for TypeNet as used to solve All-Pairs.}\label{detail_config_2}
\end{table}

\subsection{TypeNet Architecture Search} \label{allpairsresult_sup}

\begin{figure*}[!htb]
\centering
\minipage{0.74\textwidth}
  \includegraphics[width=\linewidth]{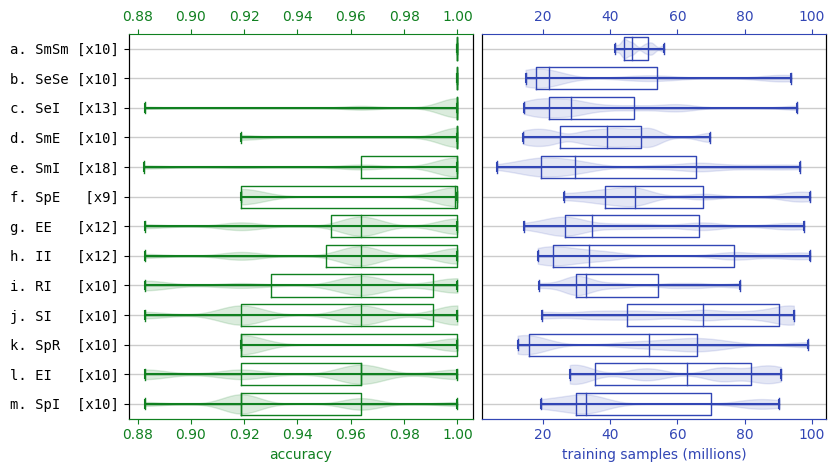}
\endminipage\hfill
\caption{4-4 All-Pairs for different activation functions, \textsc{A}$_i$.}
\label{arch_search_fig}
\end{figure*}

The main hyper-parameters and architecture-variations explored are the
feature activation, number of branches ($k$), and number of features
($n$).  First, we explored the choice of activation with $n=64$ and
$k=2$.  All activation combinations drawn from the following options
were explored and the top results are presented in Figure
\ref{arch_search_fig} : \textsc{Elu} ($\mathsf{E}$), \textsc{Identity} ($\mathsf{I}$), \textsc{Relu} ($\mathsf{R}$), \textsc{Selu} ($\mathsf{Se}$),
\textsc{Sigmoid} ($\mathsf{S}$), \textsc{SoftMax} ($\mathsf{Sm}$), \textsc{SoftPlus} ($\mathsf{Sp}$), and \textsc{Tanh} ($\mathsf{T}$).
In each figure, architectures are labeled with $n$ when $n\neq64$, and the
above abbreviations of the $k$ activations are used.  If a ``$\dashw$'' is appended,
the architecture had a wider convolution receptive field (the stride of the
third \textbf{conv} layer was 2).
\begin{equation} \label{archnotation_sup:1}
\mathsf{[\ndash] Activation_1 [... Activation_k]}[\dashw].
\end{equation}

All of the runs represented in Figure \ref{arch_search_fig} had
higher accuracy than any of the baselines.  The main conclusion from
these trials is that \textsc{SoftMax} and \textsc{Selu} are the most useful
activations.  We most frequently used \textsc{SoftMax} as the activation in
exploring the other hyper-parameters because of its low training
variance.

We studied how the number of branches, $k$, affects training; those
results are shown below with the number of training samples needed to
fully solve the 4-4 All-Pairs problem.  All trials reached 100\% accuracy,
save for one three-branch trial which got stuck at a test accuracy of
99.948\% after 30M training examples.  Based on the number of samples
needed to reach maximum test accuracy, we conclude that $k=2$ is best
for this problem.
\begin{center}
\begin{tabular}{ l l l }
branches ($k$) & accuracy  & training samples \\
\hline
 1 [$\times$9] & $1.0 \pm 0.0$                     & $57.1M \pm 3.8M$ \\
 2 [$\times$10] & $1.0 \pm 0.0$                   & $47.7M \pm 4.7M$    \\
 3 [$\times$20] & $1.0 \pm 10^{-4}$     & $49.4M \pm 8.9M$   \\
\end{tabular}\label{branches}
\end{center}
The \textsc{SoftMax} activated network with two branches was found to train
faster for more features as summarized in the following table:  \\
\begin{center}
\begin{tabular}{ l l l }
features ($n$) & accuracy  & training samples \\
\hline
 48 [$\times$9] & $1.0 \pm 0.0$       & $57.0M \pm 8.9M$ \\
 64 [$\times$10] & $1.0 \pm 0.0$     & $47.7M \pm 4.7M$    \\
 96 [$\times$20] & $1.0 \pm 0.0$     & $40.5M \pm 7.7M$.   \\
\end{tabular} \\
\end{center}
All options consistently achieved 100\% test accuracy, so this trade-off
for the 4-4 problem can be made to optimize training time or inference
time.

\subsection{More Details on the Harder All-Pairs Problems}

\begin{figure*}[!htb]
\minipage{0.23\textwidth}
  \includegraphics[width=\linewidth]{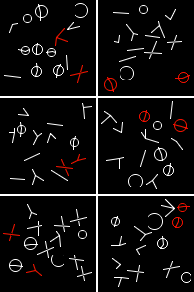}
  \endminipage\hfill
\minipage{0.74\textwidth}
  \includegraphics[width=\linewidth]{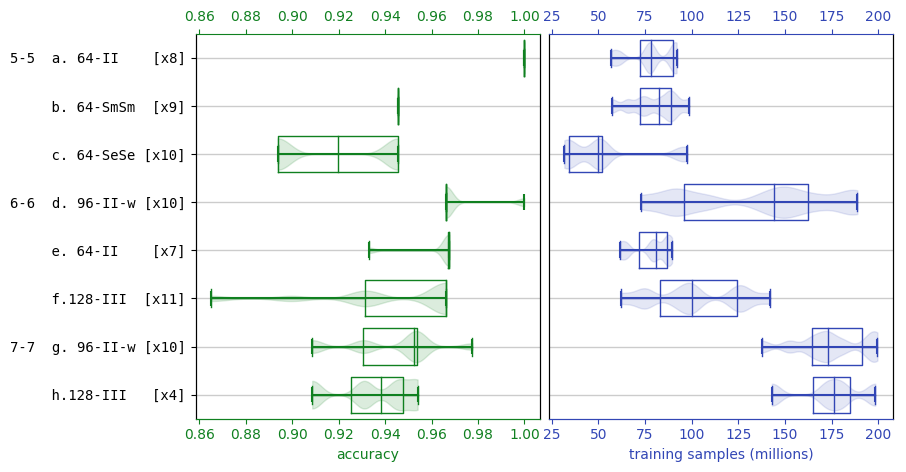}
\endminipage\hfill
\caption{\emph{Left}: Examples of incorrect test samples from TypeNet $\mathsf{96{\text -}II{\text -}w}$ trained on 7-7 All-Pairs for 200M samples.  White symbols can be paired, leaving the red symbols unpaired.
  \emph{Right}: Test results of applying TypeNet to more difficult All-Pairs problems.  Wider \textbf{conv} receptive fields are notated with ``-$\mathsf{w}$'', see text for details. }
\label{missed_sup}
\end{figure*}


The TypeNet approach cannot easily be made to solve every All-Pairs
problem; Figure \ref{missed_sup} shows results for the 5-5, 6-6, and 7-7 All-Pairs problem.
The \textsc{Identity} activation was the only activation to reach 100\% accuracy on the
5-5 and 6-6 problem, in 100\% (Fig\ref{missed_sup}-a) and 20\% (Fig\ref{missed_sup}-d) of trials respectively.  The \textsc{Selu} and \textsc{SoftMax} activation
were not successful on any of these problems in any trail within the 100M
training sample limit.

For these problems, the image size was increased
from $76\TMS76$ to $96\TMS96$ to make room for all the symbols.
This image size increase required decreasing the batch size from 600 to
400; all other training settings remained unchanged.  The large image size led us to expand the receptive field of the \textbf{conv} as notated with ``-$\mathsf{w}$'' and detailed in Section \ref{allpairsresult_sup}.  The most enlightening observations from these experiments are as follows:
\vspace{-4mm}
\begin{tightitemize}
\item The \textsc{Selu} activation (Fig\ref{missed_sup}-c) had lower accuracy
than expected from its effectiveness on the 4-4 problem.
\item On these harder problems, the \textsc{SoftMax} activation continued to
show lower variance across trials in both accuracy and training samples.
\item The $\mathsf{SmSm}$ model (Fig\ref{missed_sup}-b) consistently got stuck
at 94.6\% accuracy on the 5-5 problem, perhaps because the \textsc{SoftMax} activations are prone
to local minima.
\item The number of branches was increased to 3 and number of features to 128,
  independently and together, for the best case activations from smaller models.
  The $\mathsf{128{\text -}III}$ (Fig\ref{missed_sup}-f) model had the best test accuracy, but did worst than the simpler $\mathsf{II}$ model
  (Fig\ref{missed_sup}-e) even when trained to 200M training examples.
\item The 7-7 All-Pairs problem (Fig\ref{missed_sup}-g,h) is clearly
harder. The wider $\mathsf{96{\text -}II{\text -}w}$ (Fig\ref{missed_sup}-g) model was the best.
\item As shown in Figure \ref{missed_sup}-Left, the test samples missed by
one of the $\mathsf{96{\text -}II{\text -}w}$ models on the 7-7 problem are semantically similar: the
model incorrectly labels some samples as $true$ that have either an unpaired
\textbf{cross} and \textbf{3-star}, or an unpaired \textbf{theta} and \textbf{phi}.  For this model and trial, all of its errors fall into these two classes, though it correctly classifies some of those examples (achieving a 95\% accuracy when those two classes account for 9.5\% of the test set).  Different trials show different types of semantic errors.
\item Many variations, including mixtures of activations, more features, more branches, even wider \textbf{conv} receptive fields, and combinations of these choices, were tried to solve the 7-7 problem without success.  In the highest test accuracy observed (98\%), the misclassified images are still easy for a human to classify.
\end{tightitemize}

\subsection{Comparison to a Simple CNN}
\label{comptocnn}
\noindent
Are Lines 5 and 6 of Algorithm 1 generally useful, and do they improve the algorithm?
The table below compares (3 trials for each) the test accuracy and model size of TypeNet with a with a simple convolutional net (ConvNet) created by altering TypeNet as follows:
\begin{itemize}
\item Replace Lines 5 and 6 of Algorithm 1 with \textsc{Flatten} (passing the convolution output directly to the fully-connected layers).
\item As with larger All-Pairs images, use a stride of 2 in \textsc{Conv}$_3$.
\end{itemize}


\begin{center}
\begin{tabular}{ r | c c | c c  }
              & \multicolumn{2}{c|}{ConvNet}                        & \multicolumn{2}{c}{TypeNet}  \\
dataset  & accuracy & \# parameters & accuracy & \# parameters \\
\hline
MNIST               & 0.9953 $\pm$ 0.0002    & 2M & \textbf{0.9971 $\pm$ 0.0006} & \textbf{1M}  \\
Fashion-MNIST & \textbf{0.9409 $\pm$ 0.0005}    & 2M & 0.9346 $\pm$ 0.0011 & \textbf{1M} \\
CIFAR10           & 0.7773 $\pm$ 0.0013     & 2.5M & \textbf{0.8820 $\pm$ 0.0080} & \textbf{1M} \\
4-4 All-Pairs      & 0.8080 $\pm$ 0.0925     & 9.9M & \textbf{1.0000 $\pm$ 0.0000} & \textbf{1M} \\
\end{tabular}
\end{center}

\noindent
From this comparison, TypeNet is seen to have fewer parameters and shows significant improvements in accuracy for the hardest two datasets (CIFAR10 and 4-4 All-Pairs).  The number of parameters in TypeNet is not dependent on the input size because of the spatial summation in Line 6 of the algorithm.  We anticipate the spatial, learned histogram of TypeNet to be a useful tool in the construction of other DNN architectures.


\end{document}